\documentclass[11pt,a4paper]{article}
\usepackage[hyperref]{emnlp-ijcnlp-2019}
\usepackage{times}
\usepackage{latexsym}
\usepackage{bm}
\usepackage{amsmath}
\usepackage{amssymb}
\usepackage{multirow}
\usepackage{xspace}
\usepackage{graphicx}
\usepackage{color}
\usepackage{makecell}
\usepackage{booktabs}
\usepackage{array}
\usepackage{booktabs}
\usepackage{color,colortbl}
\usepackage{tabularx}
\usepackage{soul}
\usepackage{balance}

\newcommand{\rocstories}{ROCStories}

\usepackage{url}

\aclfinalcopy 

\newcommand{\dataName}{{{\textsc{TimeTravel}}}\xspace}
\definecolor{LightCyan}{rgb}{0.88,1,1}

\title{
Counterfactual Story Reasoning and Generation
}

\newcommand\aitwo{$^\diamondsuit$}
\newcommand\uw{$^\spadesuit$}
\newcommand\aspace{\hspace{.75em}}

\author{
  Lianhui Qin \uw\aitwo\aspace
  Antoine Bosselut \uw\aitwo\aspace
  Ari Holtzman \uw\aitwo\aspace
  Chandra Bhagavatula \aitwo\aspace \\
  \textbf{Elizabeth Clark   \uw\aspace 
  \,\,\,\,\,\,\,\,\,\,\,\,\,\,\,\,\,\,\,\,
  Yejin Choi \uw\aitwo} \\
 \uw Paul G. Allen School of Computer Science \& Engineering, University of Washington\\
 \aitwo Allen Institute for Artificial Intelligence\\
  \texttt{\text{\{lianhuiq,antoineb,ahai,eaclark7,yejin\}@cs.washington.edu}} \\
 \texttt{\text{chandrab@allenai.org}} \\
}

\date{}

\begin{document}
\maketitle
\begin{abstract}

Counterfactual reasoning 
requires predicting how alternative events, contrary to what actually happened, might have resulted in different outcomes. Despite being considered a necessary component of AI-complete systems, few resources have been developed for evaluating counterfactual reasoning in narratives.

In this paper, we propose \textit{Counterfactual Story Rewriting}: given an original story and an intervening counterfactual event, 
the task is to minimally revise the story to make it compatible with the given counterfactual event.
Solving this task will require deep understanding of causal narrative chains and counterfactual invariance, and integration of such story reasoning capabilities into conditional language generation models.

We present \dataName, a new dataset of 29,849  counterfactual rewritings, each with the original story, 
a counterfactual event, and human-generated revision of the original story compatible with the counterfactual event.
Additionally, we include 80,115 counterfactual ``branches'' without a rewritten storyline to support future work on semi- or un-supervised approaches to counterfactual story rewriting. 

Finally, we evaluate the counterfactual rewriting capacities of several competitive baselines based on pretrained language models, and assess whether common overlap and model-based automatic metrics for text generation correlate well with human scores for counterfactual rewriting.
\end{abstract}


\begin{figure}[!htb]
    \centering
    \hspace*{-2mm}
    \vspace*{-2mm}
    \includegraphics[width=.49\textwidth]{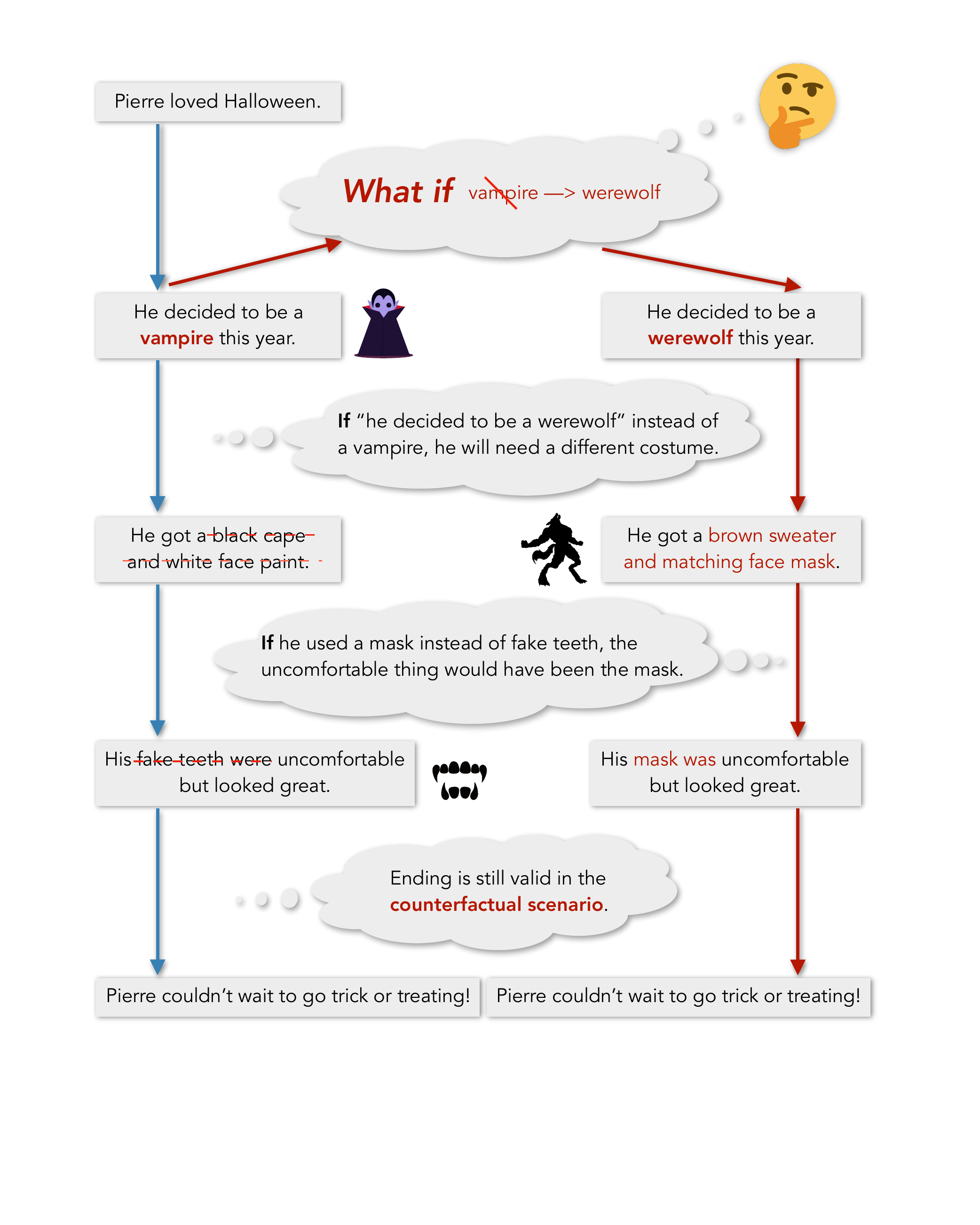}
    \caption{
    Given a short story (left column) 
    and a \emph{counterfactual context} (``He decided to be a werewolf this year''), the task is to revise the original story with minimal edits to be consistent with both the original premise (``Pierre loved Halloween'') and the new counterfactual situation. The modified parts in the new story (right column) are highlighted in \textcolor{red}{red}.
    }
    \label{fig:fig_1}
\end{figure}

 \begin{figure*}[t]
\centering
{\includegraphics[scale=0.53]{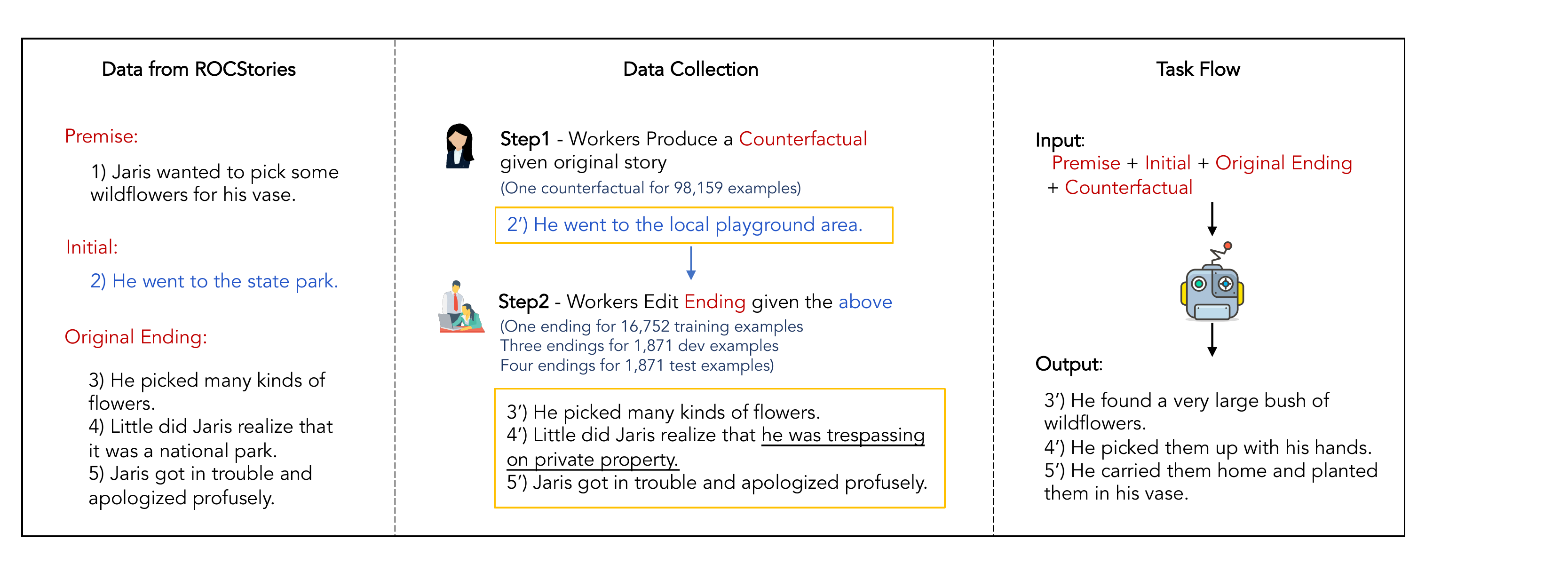}}
\caption{Data annotation process for the \dataName~dataset. Given a story from the ROCStories corpus, crowdworkers write a counterfactual sentence w.r.t the second sentence of the story. The counterfactual sentence and the original story are then presented to other workers to rewrite the story ending. Models for the task are expected to generate a rewritten ending given the original story and counterfactual sentence.}
\label{fig:fig_2} 
\end{figure*}

\section{Introduction}


A desired property of AI systems is counterfactual reasoning: the ability to predict causal changes in future events given a counterfactual condition applied to the original chain of events~\cite{goodman1947problem,bottou2013counterfactual}. 
For example, given an original story shown in the left chain in Figure~\ref{fig:fig_1}, where ``Pierre loved Halloween. He decided to be a \emph{vampire} this year. He got a \emph{black cape and white face paint}...'' and a counterfactual condition, ``what if Pierre decided to be a \emph{werewolf} instead of a \emph{vampire}?'', an intelligent system should be able to revise the subsequent events in the story appropriately, for example, that a \emph{brown sweater} would be more appropriate than a \emph{black cape}. 


This notion of counterfactuals has become increasingly relevant in several recent benchmarks such as ROC story cloze  \cite{mostafazadeh2016corpus}, COPA \cite{Roemmele2011SemEval2012T7},  and HellaSwag \cite{Zellers2019HellaSwagCA}, where the negative responses in multiple-choice problems implicitly construct counterfactual narratives. However, no existing benchmark to date has been designed to explicitly evaluate counterfactual narrative reasoning and revision as its principal focus, where a system is evaluated on its ability to make modifications to future events based on a counterfactual condition, as illustrated in Figure~\ref{fig:fig_1}. 


In this paper, we introduce \textit{Counterfactual Story Rewriting} as a new challenge to story understanding and generation. 
Given an original story and a counterfactual condition, the task is to re-write the story 
to regain narrative consistency through counterfactual reasoning.  
An important challenge in counterfactual reasoning is \emph{causal invariance}, namely, the aspects of future events that are invariant under the counterfactual conditions. This is necessary to accurately reason about the new consequences with minimal edits to the original sequence of events, instead of being confounded by spurious correlations~\cite{woodward2002mechanism,iclr2019learning}.
Therefore, a key measure of the task besides consistency is that the rewriting must perform \emph{minimal edits} to the original story. This challenges the system to reason about causal invariance, which in turn, challenges the system to reason more carefully about the causal chains of how the story unfolds.

We introduce \dataName, a new dataset with 29,849 counterfactual  revisions to support research on counterfactual narrative reasoning and revision. In addition, our dataset provides 80,115 counterfactual \emph{branches} without rewritten storylines to support potential future work on semi- or un-supervised approaches. Figure~\ref{fig:fig_2} illustrates (1) the structure of the original stories, (2) the counterfactual data construction process, and (3) the final task definition. 

We establish baseline performances of state-of-the-art neural language models on this task, such as GPT \cite{radford2018improving} and GPT-2 \cite{radford2019language}, evaluated in zero-shot, unsupervised, and supervised learning settings. 
Empirical results indicate that while these models are able to capture certain instances of counterfactual reasoning, they generally struggle with rewriting endings with full consistency. 
Our results suggest that current neural language models operate based primarily on frequent patterns in language without true understanding of the causal chains in narratives, thus requiring more focused future research to integrate reasoning capabilities in neural language models. 
\footnote{Code and data are available at \url{https://github.com/qkaren/Counterfactual-StoryRW}.} 



\section{Background}
\label{sec:background}

Counterfactual reasoning is the ability to consider \textit{alternative possibilities} that diverge from current observed narratives. 
Due to their prevalence in common reasoning situations, counterfactuals have been studied in a wide range of disciplines, 
including psychology \cite{epstude2008functional}, cognitive science \cite{byrne2002mental}, as well as natural language processing \cite{hobbs2005toward, Lawrence2018ImprovingAN, Son2017RecognizingCT}.

Meanwhile, despite the progress made in NLU tasks by adapting pretrained language representations such as BERT~\citep{devlin2018bert} or GPT \cite{radford2018improving}, models still have trouble discriminating between reasonable and unreasonable counterfactuals, as shown in \cite{Zellers2019HellaSwagCA}. Moreover, success in tasks linked to discrimination of reasonable alternatives often results in models learning to exploit latent artifacts of the dataset \cite{niven2019probing,Zellers2018SWAGAL}, rather than learning to robustly reason about counterfactuals. In response to this, we hypothesize that learning to \textit{generate the result} of counterfactual prompts will encourage models to learn to understand the underlying dynamics of a given situation, whereas discrimination between two alternatives is more likely to take advantage of dataset biases. 

This goal shares many similarities with script learning \cite{pichotta2014statistical,chambers2013event}, which attempts to canonicalize stereotypical event sequences for learning causal structure of narratives. However, because it is often difficult to capture the richness of causal dependencies with templatized structures \cite{Sap2019ATOMICAA}, we instead study counterfactual reasoning in unstructured text directly and also require the model to \emph{generate} the consequences of the counterfactual reasoning. 

The ``counterfactual event'' in our task can be viewed as a causal intervention \cite{pearl2000causality} in the latent chain of events of the story. Such interventions demand changes to the written narrative in order to abide by the shared background knowledge that human readers have about how the world works. This neatly embeds the problem of causal reasoning in a space that laymen with no knowledge of formalized causality can understand. It also allows us to evaluate the capabilities and limitations of the recent advances in neural language models in the context of counterfactual reasoning. 

Similar issues arise in the area of controllable language generation~\citep[e.g.,][]{hu2017toward}, which involves preserving the  content of text while changing it along a single or multiple dimensions, such as theme \citep{KoncelKedziorski2016ATA}, style \citep{Lample2018MultipleAttributeTR}, and sentiment \citep{shen2017style}. Reasoning in these tasks is limited to discrete axes (e.g., sentiment), which are often categorized with a closed label set (\{positive, negative\}). Because of controllability motivations, these axes and labels are generally known {\it a priori}. 
In contrast, counterfactual rewriting focuses on the causes and effects of a story, dimensions that can require more complex and diverse, yet potentially subtle, changes to accommodate the counterfactual event. Additionally, we put no restrictions on the nature of counterfactual events, yielding no clear set of discrete axes along which the story can change and no closed set of labels for them.
\section{Counterfactual Story Rewriting}
\label{sec:counterfactual}
 \begin{table*}[ht]
\centering
\begin{small}
\begin{tabular}{rp{5in}}
\toprule
\textbf{Premise} & Alec's daughter wanted more blocks to play with. \\
\textbf{Initial} & Alec figured that blocks would develop her scientific mind. \\
 \textbf{Original Ending} & \cellcolor{blue!5} Alec bought blocks with letters on them. Alec's daughter made words with them rather than structures. Alec was happy to see his daughter developing her verbal ability. \\ 
\textbf{Counterfactual} & Alec couldn't afford to buy new blocks for his daughter.\\ 
\textbf{Edited Ending} & \cellcolor{red!8} Alec decided to make blocks with letters on them instead. Alec's daughter made words with the blocks. Alec was happy to see his daughter developing her verbal ability. \\ 
\midrule
\textbf{Premise} & Ana had just had a baby girl.\\
\textbf{Initial} & She wanted her girl to have pierced ears. \\
\textbf{Original Ending} & \cellcolor{blue!5} She took her baby to the studio and had her ears pierced. Then she fastened tiny diamond studs into the piercings. Ana loved the earrings. \\ 
\textbf{Counterfactual} & She didn't like the idea of having her ears pierced.\\
 \textbf{Edited Ending} & \cellcolor{red!8} She decided not to take her baby to the studio to get her ears pierced. So she took tiny diamond stickers and stuck them to her ear. Ana loved the fake earrings. \\ 
\bottomrule
\end{tabular}
\end{small}
\caption{Examples from \dataName }
\label{tab:dataset_examples}
\end{table*}

\subsection{Task}

\label{ssec:counterfactual:task}
We now formally introduce the task and establish the notation used in the paper. Each example consists of a five-sentence story $S=(\bm{s}_1, \dots, \bm{s}_5)$ with a general structure where the first sentence $\bm{s}_1$ sets up the \emph{premise}, the second sentence $\bm{s}_2$ provides more information of the \emph{initial context}, and the last three sentences $\bm{s}_{3:5}$ are the \emph{original ending} of story. We are further given an additional sentence $\bm{s}'_2$, which is counterfactual to the initial context $\bm{s}_2$. That is, $\bm{s}'_2$ states something contrary to that in $\bm{s}_2$, which in turn can make the original ending $\bm{s}_{3:5}$ no longer valid. Thus, the goal of the task is to rewrite the ending, such that the \emph{edited ending} $\bm{s}_{3:5}'$ minimally modifies the original one and regains narrative coherency to the new counterfactual context.

The minimum edit goal differentiates our task from previous story ending studies, which have mostly focused on consistency in a given context. To achieve consistency with minimal edits, a model must understand the key mechanisms that drive the story's narrative so that it can filter out spurious correlations and capture counterfactual invariance. We thus consider the new task as a suitable testbed for studying counterfactual reasoning in combination with language generation.


\subsection{Dataset: \dataName}

Our dataset is built on top of the ROCStories corpus \cite{mostafazadeh2016corpus}, which contains 98,159 five-sentences stories in the training set, along with 3,742 stories in the evaluation sets. Each story was written by crowdworkers. To collect counterfactual events and new story continuations for \dataName, we employ workers from Amazon Mechanical Turk (AMT) for a two-step task, which we describe in detail below. 

\subsection{Data Collection}

 \paragraph{Counterfactual Event Collection}  We present workers with an original five-sentence story $S=(\bm{s}_1, \bm{s}_2,\dots, \bm{s}_5)$ and ask them to produce a counterfactual event $\bm{s}'_2$ based on $\bm{s}_2$. Workers are instructed to produce counterfactual sentences $\bm{s}'_2$ that are: \\
    (1) Topically related to the original context sentence $\bm{s}_2$, rather than a completely new sentence. \\
    (2) Relevant to the original premise sentence, $\bm{s}_1$, allowing for a coherent story continuation. \\
    (3) Influential to the subsequent storyline, such that at least one of the original ending's sentences, $\{ \bm{s}_3$, $\bm{s}_4$, $\bm{s}_5 \}$ is no longer appropriate given $\bm{s}_1$ and $\bm{s}'_2$, necessitating a rewritten story ending. 

 \paragraph{Continuation Rewriting} Once a counterfactual sentence $\bm{s}'_2$ is provided, we present it to a new set of workers with the original story $S=(\bm{s}_1, \bm{s}_2,\dots, \bm{s}_5)$. Now that $\bm{s}'_2$ invalidates the original storyline, workers are instructed to make minimal edits to $\bm{s}_{3:5}$, such that the narrative is coherent again. Before beginning, workers are instructed to validate whether the counterfactual event satisfies the requirements from the previous stage of the pipeline. If not, we ask them to rewrite the counterfactual again, and the continuation rewriting step is reassigned to a new worker. 


\paragraph{Summary} We provide examples from the \dataName dataset in  Table~\ref{tab:dataset_examples} and summarize its scale in Table~\ref{tab:stats}. Overall, we collect 16,752 training examples of a counterfactual context and a rewritten ending. We also collect an additional 80,115 counterfactual contexts for the training set with no rewritten ending to support future work in unsupervised learning on this task. For the development and test sets, we gather multiple counterfactual contexts and rewritten endings for \textit{each} example (3 new endings for development and 4 for test). Information regarding quality control and cost are provided in Appendix~\ref{app:data}.

\begin{table}
\small
    \centering
    \begin{tabular}{lrrr}
    \cmidrule[\heavyrulewidth]{1-4}
      & Train & Valid & Test \\ 
    \cmidrule{1-4}
    \multicolumn{3}{l}{\it ROCStories data:}  \\
    \# Stories     & 98,159   & 1,871   & 1,871 \\ 
    \cmidrule{1-4}
    \multicolumn{3}{l}{\it \dataName :}  \\
    \# Counterfactual Context    & 96,867   & 5,613   & 7,484 \\ 
    \# Edited Ending   & 16,752    & 5,613   & 7,484 \\
    \cmidrule[\heavyrulewidth]{1-4}
    \end{tabular}
    \caption{Dataset statistics} 
    \label{tab:stats}
\end{table}

\label{ssec:counterfactual:dataset}

\section{Learning a Counterfactual Rewriter}
\label{sec:model}

Recent work in constructing large-scale generative language models based on transformers \cite{radford2018improving,radford2019language} has led to considerable improvements in natural language generation tasks. Due to their current prominence, we use them as baselines to study the extent to which the current neural text generation systems can perform and fail counterfactual narrative reasoning and revision. 
We focus on the family of GPT models, including GPT~\cite{radford2018improving} and the latest small- (GPT2-S) and medium-sized (GPT2-M) transformer models from \citet{radford2019language}. For each of the three pretrained language models, we fine-tune with multiple objectives, leading to 14 different model variants for the task, which we describe in more detail below. 

\subsection{Unsupervised Training}

Constructing large-scale counterfactual revision dataset is costly. Therefore, an ideal system must learn to reason without direct supervision. Toward this goal, we examine how unsupervised approaches to counterfactual story rewriting perform on our evaluation task. We devise the following unsupervised settings for models to learn to generate counterfactual story endings.

\paragraph{Zero-shot (ZS)}
In our simplest setting, we evaluate the counterfactual reasoning abilities already learned by these models due to pretraining on large corpora: the BooksCorpus dataset \cite{Zhu2015AligningBA} for GPT and the WebText corpus for GPT-2 \cite{radford2019language}. In this setting, models are not trained on any portion of the training data from \dataName~and must instead produce counterfactual rewritten stories for the evaluation set using only the representations learned from pretraining. At test time, the model receives the premise and the counterfactual context $(\bm{s}_1, \bm{s}_2')$  as input and generates the tokens that constitute the rewritten counterfactual outcome.

\paragraph{Fine-tuning (FT)} 
Because the domains on which both the GPT and GPT2 models were trained are broad and more complex than the domain of ROCStories, we investigate whether adapting the language model to the data distribution of ROCStories is helpful for learning to reason about counterfactuals. In this setting, the model is further fine-tuned to maximize the log-likelihood of the stories in the ROCStories corpus:
\vspace*{-4mm}
\begin{equation}
\begin{split}
\mathcal{L}^{ft}(\bm{\theta}) = \log p_\theta(S),
\end{split}
\label{eq:loss-fineune}
\end{equation}
where $p_\theta$ is the language model with parameters $\bm{\theta}$, and $S$ is the original story as defined in Section~\ref{ssec:counterfactual:task}. This fine-tuning step encourages the model to generate text with the same consistent style of the stories. Similar to the zero-shot setting, the premise and the counterfactual sentence $(\bm{s}_1, \bm{s}_2')$ are provided as input to the model.

\paragraph{Fine-tuning + Counterfactual (FT + CF)}
The above training loss, however, does not make use of the additional 81,407 counterfactual training sentences for fine-tuning. To inform the model with a larger set of possible counterfactual narratives in the training data, we propose an additional loss function that fits the model to the counterfactual sentences given the premise sentence:

\vspace*{-6mm}
\begin{equation}
\begin{split}
\mathcal{L}^{cf}(\bm{\theta}) = \log p_\theta(\bm{s}_2' | \bm{s}_1),
\end{split}
\label{eq:loss-counterfact}
\end{equation}
where $p_\theta(\bm{s}_2' | \bm{s}_1)$ denotes that the language model first reads the premise $\bm{s}_1$ and maximizes the log-likelihood of counterfactual sentence $\bm{s}_2'$.
The model is fine-tuned with both objectives in Eqs~\eqref{eq:loss-fineune} and \eqref{eq:loss-counterfact}:
\begin{equation}
\mathcal{L}^{ft+cf}(\bm{\theta}) = \mathcal{L}^{ft} + \mathcal{L}^{cf},
\end{equation}
%
and receives inputs in the same format as the zero-shot and fine-tuned models at test time.
\begin{table*}[ht]
\centering
\begin{small}
\begin{tabular}{r p{5in}}
\toprule

Premise & Rob loved to play soccer and was on his middle school team. \\ 
Initial & They had just started their season and were playing their first game. \\ 
Original Ending & Rob accidentally tripped over another player during the game. The other player was mad, but Rob apologized and kept his composure. Rob ended up playing the best game of his life! \\ 
Counterfactual & When he graduated from middle school, he started learning to play the guitar. \\ 
\\
\rowcolor[gray]{0.95} \multicolumn{1}{c}{\textbf{Model}} & \multicolumn{1}{c}{ \textbf{Generated Ending}} \\ 
GPT-M + ZS & He played in bands for about a year, then got into touring with a band called The Muffs. He wrote the music for a few of The Muffs' songs. He was a very talented singer-songwriter and guitarist. \\ 
GPT-M + FT + CF & He took classes on the guitar for years to get better. He took a solo lesson and had a great time. Rob now is a famous guitarist. \\ 
GPT-M + Supervised & He accidentally tripped over the guitar during the song. The other person was mad, but Rob apologized and kept his composure. Rob ended up playing the best song of his life. \\ 
\midrule
Premise & I called Carly to ask her on a date. \\ 
Initial & I got her voicemail. \\
Original Ending & I texted her that we should talk soon. She texted me back, asking why we should talk soon. I told her I liked her. \\
Counterfactual & She said that we should just stay friends which made me really sad. \\ 
\\
\rowcolor[gray]{0.95} \multicolumn{1}{c}{\textbf{Model}} & \multicolumn{1}{c}{ \textbf{Generated Ending}} \\ 
GPT-M + ZS & She said that she went out and did something really stupid and that was all I could think about. I did my best to be nice to Carly by telling her how much I loved her and that I would be her best friend no matter what. There were nights when I felt like crying because I couldn't get out of bed. \\ 
GPT-M + FT + CF & I decided to watch a movie instead. After that I decided I should see a movie I had never seen before. I went to the theater and bought a ticket. \\ 
GPT-M + Sup & I texted her that we should just stay friends soon. She texted me back, asking why we should just stay friends soon. I told her I liked her. \\ 

\bottomrule
\end{tabular}
\end{small}
\caption{Model Outputs}
\label{tab:examples}
\end{table*}

\paragraph{Reconstruction + Counterfactual (RC + CF)}
One issue with the above training procedures is that models are not explicitly trained to retain as much text of the original outcome $\bm{x}_{3:5}$ as possible (i.e., minimum edits). If these models are to learn to ``rewrite" the original story ending given the counterfactual sentence, rather than learning to produce a completely new plot, they must be able to condition on the original ending during generation. 
Motivated by this requirement and following the goal of developing unsupervised methods for counterfactual rewriting, we design a reconstruction objective for learning a noisy reproduction of the original ending. Specifically, we provide the model with both the original story and a masked context as input $(S, [s], \bm{s}_1, [mask])$ and train the model to reconstruct the original ending $\bm{s}_{3:5}$:
\begin{equation}
\begin{split}
\mathcal{L}^{rc}(\bm{\theta}) = \log p_\theta(\bm{s}_{3:5} | S, [s], \bm{s}_1, [mask]),
\end{split}
\label{eq:loss-recon}
\end{equation}
where  \texttt{[s]} denotes a separator token and \texttt{[mask]} is a special mask token. In this setting, the model first reads the original story $S$ followed by the separation token \texttt{[s]}, and then reads the premise $\bm{x}_1$ again, followed by the mask token \texttt{[mask]}, which serves as a placeholder sentence for the counterfactual sentence. This objective encourages the model to reproduce the original ending $\bm{s}_{3:5}$ in the general case where the second sentence is not specified, thereby encouraging generations similar to the original ending regardless of the counterfactual provided. At test time, we replace \texttt{[mask]} in the input with the counterfactual sentence $\bm{s}_2'$, and the model must generate the continuation of $(S, [s], \bm{s}_1, \bm{s}_2')$. 
We also use the objective from Eq~\eqref{eq:loss-counterfact} above to inform the model with counterfactual information during training.

\subsection{Supervised Training (Sup)}
Our dataset also provides 16,752 training instances that include human annotated rewritten endings for supervised learning. To assess whether being able to train directly on alternative endings is helpful for learning counterfactual narrative understanding, we train models on this portion of data in a supervised manner. More concretely, the input to the model contains the full information $(S, [s], \bm{s}_1, \bm{s}_2')$, and we train the model to maximize the log-likelihood of ground-truth rewritten endings:
\begin{equation}
\begin{split}
\mathcal{L}^{s}(\bm{\theta}) = \log p_\theta(\bm{s}_{3:5}' | S, [s], \bm{s}_1, \bm{s}_2').
\end{split}
\label{eq:loss-super}
\end{equation}

\noindent where \texttt{[s]} denotes a separator token.


\subsection{Hyperparameters}
We largely follow the same training and inference setups as in \citet{radford2018improving} for the GPT model and \citet{radford2019language} for the GPT2 variants. Experiments are implemented with the text generation toolkit Texar~\citep{hu2019texar}. We provide more details in Appendix~\ref{app:train}. 



\section{Human Study of Rewritten Sentences}
\label{sec:human}

To assess the quality of rewritten endings, we conduct two sets of human evaluation. To give a sense of the model generation, Table~\ref{tab:examples} presents example outputs by a subset of representative models on two test cases.
\begin{table}[t]
\small
\centering
\begin{tabular}{lrrr}
\toprule
Model & \textbf{Pre (1)} & \textbf{Plot (2)} & \textbf{CF (3)} \\
\toprule
\small{GPT + ZS}            & 1.945          & 1.290         & 1.555          \\ 
\small{GPT2-S + ZS}         & 1.945          & 1.335         & 1.475          \\ 
\small{GPT2-M + ZS}         & 2.435          & 1.615         & 2.045          \\ 
\midrule
\small{GPT + FT}            & 2.485          & 1.750         & 2.005          \\ 
\small{GPT2-S + FT}         & 2.365          & 1.645         & 1.895          \\ 
\small{GPT2-M + FT}         & 2.580          & 1.790         & \textbf{2.070} \\ 
\midrule
\small{GPT + FT + CF}       & 2.310          & 1.595         & 1.925          \\ 
\small{GPT2-S + FT + CF}    & 2.310          & 1.640         & 1.850          \\ 
\small{GPT2-M + FT + CF}    & 2.395          & 1.650         & 1.945          \\ 
\midrule
\small{GPT2-S + RC + CF}    & 2.240          & 2.090         & 1.500          \\ 
\small{GPT2-M + RC + CF}    & \textbf{2.780} & 2.595         & 1.660          \\ 
\midrule
\small{GPT + Sup}       & 2.630          & \textbf{2.690}& 1.460          \\ 
\small{GPT2-S + Sup}        & 2.705          & 2.650         & 1.625          \\ 
\small{GPT2-M + Sup}        & 2.750          & 2.620         & 1.820         \\
\midrule 
Human               & 2.830          & 2.545         & 2.520         \\
\bottomrule
\end{tabular}
\caption{Likert scale scores for different models. The top performing model for each question is \textbf{bolded}.}
\label{tab:likert}
\end{table}

\subsection{Rewritten Sentence Scoring}
\label{ssec:human:1}


\paragraph{Setup} 
In this setting, workers from Amazon Mechanical Turk were presented 100 outputs from 14 different models. For each example, two workers were presented the original premise sentence, the original ending, the counterfactual sentence, and the rewritten ending, and asked to answer the following three questions 
 on a 3-point Likert scale: \\
\textbf{(1)} Does the rewritten ending keep in mind details of the original premise sentence? \\
\textbf{(2)} Is the plot of the rewritten ending relevant to the plot of the original ending? \\
\textbf{(3)} Does the rewritten ending respect the changes induced by the counterfactual sentence? \\
In addition to evaluating the 14 models, we also provided gold human annotated counterfactual endings for the same 100 test examples to compute an expected upper bound for how models should perform. We present the results from this study in Table~\ref{tab:likert} and share key observations below. \footnote{The average Krippendorff alpha for all three questions is 0.42 ("moderate"). \cite{ageeva2015evaluating})} 

\paragraph{Model Size and Pretraining Data} 
We observe that models with more parameters are better at the counterfactual rewriting task than smaller models. The GPT2-M variants consistently outperform the GPT and GPT2-S models, regardless of the objective on which the model was trained. Interestingly, however, the GPT model appears to generally outperform the GPT2-S model on the counterfactual question \textbf{(3)}, indicating that the domain on which models are pretrained does affect how adaptable their representations are to the story rewriting task.

\begin{table}[t]
\small
\centering
\begin{tabular}{@{}ll|c|ll@{}}
\toprule
\multicolumn{5}{c}{COUNTERFACTUAL - Human Judges Preferred}                              \\ \midrule
\rowcolor[gray]{0.95} \multicolumn{2}{c|}{Best model} & \multicolumn{1}{c|}{Neutral} & \multicolumn{2}{c}{Comparator} \\
M+Sup & 20.0 & 7.0  & \textbf{29.5} & M+FT+CF         \\
M+Sup & 19.0 & 3.0  & \textbf{38.5} & M+FT            \\
M+Sup & \textbf{23.5} & 14.0 & 4.5  & M+Recon+ CF     \\
M+Sup & 26.5 & 5.0  & \textbf{33.5} & M+ zero-shot    \\
M+Sup & \textbf{14.0} & 18.5 & 6.0  & S+Sup     \\
M+Sup & \textbf{18.5} & 20.0 & 8.0  & GPT + Sup \\ 
\midrule
M+Sup & 10.0 & 15.0 & \textbf{52.0} & Human            \\
 \bottomrule
\vspace{0.2mm}
\end{tabular}
\\
\begin{tabular}{@{}ll|c|ll@{}}
\toprule
\multicolumn{5}{c}{PLOT - Human Judges Preferred}                              \\ \midrule
\rowcolor[gray]{0.95} \multicolumn{2}{c|}{Best model} & \multicolumn{1}{c|}{Neutral} & \multicolumn{2}{c}{Comparator}  \\
M+Sup & \textbf{57.5} & 14.5 & 13.5 & M+FT+CF        \\
M+Sup & \textbf{58.5} & 16.5 & 12.5 & M+FT           \\
M+Sup & 11.5 & 60.0 & \textbf{16.5} & M+Recon+CF     \\
M+Sup & \textbf{63.0} & 14.5 & 11.0 & M+zero-shot    \\
M+Sup & 11.5 & 62.5 & \textbf{12.5} & S+Sup    \\
M+Sup & 14.5 & 61.0 & \textbf{15.0} & GPT+Sup  \\
\midrule
M+Sup & 22.0 & 47.5 & \textbf{25.0} & Human         \\
\bottomrule
\vspace{0.2mm}
\end{tabular}
\\
\begin{tabular}{@{}ll|c|ll@{}}
\toprule
\multicolumn{5}{c}{PREMISE - Human Judges Preferred}                              \\ \midrule
\rowcolor[gray]{0.95} \multicolumn{2}{c|}{Best model} & \multicolumn{1}{c|}{Neutral} & \multicolumn{2}{c}{Comparator} \\
M+Sup & \textbf{35.5} & 31.0 & 16.5 & M+FT+CF          \\
M+Sup & \textbf{32.5} & 39.5 & 14.0 & M+FT           \\
M+Sup & \textbf{10.5} & 65.0 & 9.0  & M+Recon+CF     \\
M+Sup & \textbf{46.5} & 29.5 & 13.0 & M+zero-shot     \\
M+Sup & \textbf{8.5}  & 71.0 & 7.5  & S+Sup    \\
M+Sup & \textbf{12.0} & 68.0 & 7.5  & GPT+Sup \\
\midrule
M+Sup & 12.5 & 59.0 & \textbf{22.5} & Human  \\
 \bottomrule
\end{tabular}
\caption{Pairwise human comparison between the best model (GPT2-M + Sup) and comparison models on all three questions. ``Neutral'' means both are ``equally good''. Percentage of ``equally bad'' are omitted.}
\label{tab:results}
\end{table}

\paragraph{Domain Adaptation}
Another pattern we notice is that fine-tuning on the ROCStories data (FT) is always helpful for increasing performance on counterfactual relevance (\textbf{CF} \textbf{(3)} in Table~\ref{tab:likert}), indicating adapting to the ROCStories-style language distribution helps the model learn to produce relevant rewrites for counterfactuals, especially for models with fewer parameters. The \textbf{Plot (2)} question in Table~\ref{tab:likert} indicates why this might be the case, as the zero-shot models tend to produce more creative rewritings that are not at all tied to the original story. Interestingly, however, fine-tuning with the larger set of counterfactuals (CF loss) does not seem to help in rewriting endings that relate to the counterfactuals well.

\paragraph{Supervised vs. Unsupervised Learning} A surprising observation is that using the dataset of labeled rewritten endings for training does not seem to help the language models learn to rewrite endings better. While the supervised models are generally able to adhere to the plot better than unsupervised methods, their new endings do not score well on question \textbf{(3)}, indicating that they may be copying the original ending or learning to paraphrase the original story ending without acknowledging the counterfactual sentence. This points to the fact that this task cannot be trivially solved by adding more paired data, since adding more data merely simplifies to having more stories in the dataset, without necessarily learning to handle counterfactuals more effectively.
\subsection{Pairwise Model Preference}
\label{ssec:human:2}

\paragraph{Setup}
We conduct a pairwise comparison between the best model (GPT2-M + Sup) with other models along the same three dimensions as in the first evaluation setting (section~\ref{ssec:human:1}). Specifically, crowdworkers were presented outputs of a pair of systems, and asked to choose which one is better, or ``equally good'' or ``equally bad'', in terms of each of the three criteria. 
As in section~\ref{ssec:human:1}, we evaluate 100 outputs of each model.

\paragraph{Results}
In Table~\ref{tab:results}, we present the human preference results, showing that the best model outperforms the comparison baselines in terms of consistency with premise, while being less consistently better with regards to the other two questions. Interestingly, a model that performs better on one of the evaluated dimensions often performs worse for another question, indicating plenty of room for future work in 
counterfactual reasoning for story rewriting.

\section{Challenges for Automatic Metrics}
%
To provide further insight into the performance of candidate models, we explore how different automatic metrics evaluate the produced generations.

\begin{table}[t]
    \centering
     \small
    \begin{tabular}{l|r|r|r}
        \toprule
        \multicolumn{1}{l}{Metric} & 
        \multicolumn{1}{c}{\textbf{(1) Prem}} & 
        \multicolumn{1}{c}{\textbf{(2) Plot}} & 
        \multicolumn{1}{c}{\textbf{(3) CF}}  \\
        \toprule
        BLEU-4  & \cellcolor{blue!10} \textbf{.2623} 
                & \cellcolor{blue!10}\textbf{.6792} 
                & \cellcolor{red!15}\textbf{-.1804}\\
        ROUGE-L & \cellcolor{blue!10} \textbf{.3187} 
                & \cellcolor{blue!10}\textbf{.7484} 
                & \cellcolor{red!15}\textbf{-.1423}\\
        WMS & \cellcolor{blue!10}\textbf{.2713} 
            & \cellcolor{blue!10}\textbf{.5809} 
            & \cellcolor{red!15}-.0343 \\
        S+WMS &  {\cellcolor{blue!10} \textbf{.2789}} 
              & \cellcolor{blue!10}\textbf{.6075} 
              & \cellcolor{red!15}-.0538\\
        BERT    & \cellcolor{blue!10} \textbf{.2124} 
                & \cellcolor{blue!10}\textbf{.1929} 
                & \cellcolor{blue!10} \textbf{.1067}\\
        BERT-FT & \cellcolor{blue!10} \textbf{.2408} 
                & \cellcolor{blue!10}\textbf{.1847} 
                & \cellcolor{blue!10} \textbf{.0995}\\

    \end{tabular}
    \caption{Pearson correlation between automatic metrics and human scores. \textbf{Bolded} numbers are statistically significant at p $<$ 0.05.}
    \label{tab:correlation}
\end{table}

\subsection{Metrics}
\paragraph{Overlap Metrics}
The most common metrics used in evaluating text generation are based on textual overlap between a candidate generated sequence and set of reference sequences provided by the dataset. \textbf{BLEU} \cite{papineni2002bleu} is perhaps the most widely used metric in text generation, which computes the number of overlapping $n$-grams between the generated and reference sequences. 
Another commonly used metric in text generation (though originally designed for extractive summarization) is \textbf{ROUGE-L} \cite{lin2004rouge}, which measures the length of the longest common subsequence (LCS) between a candidate generation and reference. We report the performance of all models on both of these metrics.

\begin{table*}
\small
    \centering
    \begin{tabular}{lrrrrrr}
    \cmidrule[\heavyrulewidth]{1-7}
      & BLEU-4 & ROUGE-L & BERT &  BERT-FT & WMS & W+SMS  \\ 
    \cmidrule[\heavyrulewidth]{1-7 }
    \rowcolor[gray]{0.95} \multicolumn{4}{c}{\it Training: Pretrained Only} &  \multicolumn{3}{c}{\it Input: $\bm{s}_1 \bm{s}'_2$}  \\
    GPT + zero-shot         & 1.25      & 18.26      & 59.50 &   58.28  &    0.30    & 0.97 \\
    GPT2-S + zero-shot      & 1.28      & 20.27      & 59.62 &   58.11  &    0.33    & 1.09 \\
    GPT2-M + zero-shot      & 1.51      & 19.41      & 60.17 &   58.59  &    0.34    & 1.12\\
     \rowcolor[gray]{0.95} \multicolumn{4}{c}{\it Training: Unsupervised + Generative} &  \multicolumn{3}{c}{\it Input: $\bm{s}_1 \bm{s}'_2$} \\ 
    GPT + FT                & 4.20      & 24.55      & 64.38 &  62.60    &  0.56  & 1.48 \\
    GPT2-S + FT             & 3.78      & 24.18      & 64.25 &  62.60    &  0.54  & 1.40\\
    GPT2-M + FT             & 4.09      & 24.08      & 62.23 &  62.49    &  0.53  & 1.42\\
    GPT + FT + CF           & 3.82      & 24.21      & 64.48 &  62.66  &  0.57  & 1.45\\
    GPT2-S + FT + CF        & 3.96      & 24.06      & 64.50 &  62.71   &  0.53  & 1.44 \\
    GPT2-M + FT + CF        & 4.00      & 24.38      & 64.31 &  62.59  & 0.48   & 1.33  \\
    \rowcolor[gray]{0.95} \multicolumn{4}{c}{\it Training: Unsupervised + Discriminative} &  \multicolumn{3}{c}{\it Input: $\bm{s}_1\bm{s}_2\bm{y}[S]\bm{s}_1[MASK]$}  \\
    GPT2-S + Recon + CF     & 47.08      & 51.19     & \textbf{63.82} &   62.36  &  5.53  & 8.08 \\
    GPT2-M + Recon + CF     & \textbf{76.57}      & \textbf{71.35}      & 64.15 &   62.49  &  \textbf{18.29}  & \textbf{20.87} \\
    \cmidrule{1-7}
    \rowcolor[gray]{0.95} \multicolumn{4}{c}{\it Training: Supervised + Discriminative} & \multicolumn{3}{c}{\it Input:  $\bm{s}_1\bm{s}_2\bm{y}[S]\bm{s}_1\bm{s}'_2$}  \\
    GPT + Sup      & 80.09      & 75.03      & 64.15  &   62.36  & 20.93    & 23.37 \\
    GPT2-S + Sup     & 79.03      & 73.31      & 64.14  &   62.40  & 20.57    & 22.97\\
    GPT2-M + Sup    & 76.63      & 74.42      & 64.06  &   \textbf{62.33}  & 19.62    & 22.01\\
    \cmidrule{1-7}
    Human                   & 65.12      & 68.58      & 63.58  &  61.82   & 16.95   & 19.16\\
    \cmidrule[\heavyrulewidth]{1-7}
    \end{tabular}
    \caption{Results on automatic metrics for the cross-product of the models and loss functions proposed in Section~\ref{sec:model}. \textbf{Bolded} results are closest to the human score.} 
    \label{tab:model_results}
\end{table*}

\paragraph{Model-based Metrics} Although BLEU and ROUGE are widely used in text generation, they use exact string matching, and thus fail to robustly match paraphrases and capture semantically-critical ordering changes. Recently, there has been a growing body of work in producing model-based metrics \cite{Lowe2017TowardsAA} that use trained models and embeddings to score a sequence. 

\citet{Kusner2015FromWE}
proposed Word Mover's Distance, which defines the distance between two texts as the minimal cost of transforming one sequence's word embeddings to the other's. 
The measure finds a matching between the two texts that minimizes the total Euclidean distance between the matched word embeddings.
Following \citet{Kilickaya2017ReevaluatingAM}, we take the negative exponential of this distance to get \textbf{Word Mover's Similarity (WMS)}.
%
More recently, \citet{ClarkEtAl2019} proposed \textbf{Sentence + Word Mover's Similarity (S+WMS)} to extend WMS for longer multi-sentence texts by using sentence representations in the minimum distance calculation in addition to word embeddings.\footnote{We follow previous work and use GloVe embeddings \cite{Pennington2014GloveGV} to represent words and the averaged word embeddings to represent sentences.}

Other recent methods use contextualized embeddings \cite{devlin2018bert} to compute similarity between sequences. We use \textbf{BERTScore} \cite{zhang2019bertscore}, which computes cosine similarity between two sentences using BERT encodings. \citeauthor{zhang2019bertscore} show that BERTScore correlates better with human judgments than existing metrics such as BLEU, ROUGE, and other learning-based metrics. 
To adapt BERTScore to our task, we finetune BERT on \rocstories~using the same training framework from \citet{devlin2018bert} and compute \textbf{BERT-FT} the same way as before. 
%


\subsection{Human Correlation with Metrics}

Recent work in text generation \cite{Wiseman2017ChallengesID} and dialogue \cite{Liu2016HowNT} have explored the limitations of automatic metrics for text production tasks. Due to the highly semantic nature of the counterfactual rewriting task and the need to recognize subtle changes in event descriptions, we anticipate that automatic metrics would have difficulty assessing rewritten endings. 
To test the correlation between available evaluation metrics for long-form generation and human opinions of quality of counterfactual generations, we compute the Pearson Correlation between automatic scores and human scores for 800 validation set data points, 300 taken from the gold annotations and 100 generated from each of the 5 GPT2-M variants.\footnote{We include both human annotations and model-generated outputs in this computation to encourage diversity of source.} For each example, we use the same questions and Likert scale evaluation as in \S\ref{sec:human} and report the results in Table~\ref{tab:correlation}.

As expected, the automatic metrics are decently correlated with human scores for adherence to the premise sentence and plot. However, these same metrics correlate negatively with question \textbf{(3)} -- adherence to the counterfactual sentence -- indicating poor measurement of counterfactual understanding if they were to be reported in their typical manner (i.e., higher score indicating superior performance). Only the BERTScore metrics appear to positively correlate with human scores for counterfactual understanding, making them usable for evaluating generations across properties related to all three questions. However, the correlation is weak, and the results in Table~\ref{tab:model_results} indicate that the BERTScore metrics are difficult to distinguish between models.



\section{Conclusion}
\label{sec:conclusion}

We introduced a new task of \textit{Counterfactual Story Rewriting} that challenges current language understanding and generation systems with counterfactual reasoning. 
Our new dataset, \dataName, provides nearly 30k counterfactual revisions to simple commonsense stories together with over 100k counterfactual sentences. We establish baseline performances of state-of-the-art neural language models  with over 14 model variants with zero-shot, unsupervised and supervised settings. The empirical results demonstrate that while neural language models show promises, they generally have difficulties in rewriting the consequences of the counterfactual condition with full consistency, suggesting more focused research on integrating true reasoning capabilities to neural language models.


\section*{Acknowledgements}
We thanks the anonymous reviewers as well as Michel Galley, Jianfeng Gao, and others for many helpful comments. 
This research was supported in part by NSF (IIS-1524371), DARPA CwC through ARO (W911NF15-1-0543), DARPA MCS program through NIWC Pacific (N66001-19-2-4031), and Allen Institute for AI.

\newpage
\balance
\bibliography{emnlp-ijcnlp-2019}

\begin{thebibliography}{38}
\expandafter\ifx\csname natexlab\endcsname\relax\def\natexlab#1{#1}\fi

\bibitem[{Ageeva et~al.(2015)Ageeva, Forcada, Tyers, and
  P{\'e}rez-Ortiz}]{ageeva2015evaluating}
Ekaterina Ageeva, Mikel~L Forcada, Francis~M Tyers, and Juan~Antonio
  P{\'e}rez-Ortiz. 2015.
\newblock Evaluating machine translation for assimilation via a gap-filling
  task.
\newblock In \emph{Proceedings of the 18th Annual Conference of the European
  Association for Machine Translation}, pages 137--144.

\bibitem[{Bottou(2019)}]{iclr2019learning}
Leon Bottou. 2019.
\newblock Learning representations using causal invariance.
\newblock In \emph{ICLR}.

\bibitem[{Bottou et~al.(2013)Bottou, Peters, Qui{\~n}onero-Candela, Charles,
  Chickering, Portugaly, Ray, Simard, and Snelson}]{bottou2013counterfactual}
L{\'e}on Bottou, Jonas Peters, Joaquin Qui{\~n}onero-Candela, Denis~X Charles,
  D~Max Chickering, Elon Portugaly, Dipankar Ray, Patrice Simard, and
  Ed~Snelson. 2013.
\newblock Counterfactual reasoning and learning systems: The example of
  computational advertising.
\newblock \emph{JMLR}.

\bibitem[{Byrne(2002)}]{byrne2002mental}
Ruth~MJ Byrne. 2002.
\newblock Mental models and counterfactual thoughts about what might have been.
\newblock \emph{Trends in cognitive sciences}, 6(10):426--431.

\bibitem[{Chambers(2013)}]{chambers2013event}
Nathanael Chambers. 2013.
\newblock Event schema induction with a probabilistic entity-driven model.
\newblock In \emph{EMNLP}, pages 1797--1807.

\bibitem[{Clark et~al.(2019)Clark, Celikyilmaz, and Smith}]{ClarkEtAl2019}
Elizabeth Clark, Asli Celikyilmaz, and Noah~A. Smith. 2019.
\newblock Sentence mover's similarity: Automatic evaluation for multi-sentence
  texts.
\newblock In \emph{ACL}.

\bibitem[{Devlin et~al.(2018)Devlin, Chang, Lee, and
  Toutanova}]{devlin2018bert}
Jacob Devlin, Ming-Wei Chang, Kenton Lee, and Kristina Toutanova. 2018.
\newblock Bert: Pre-training of deep bidirectional transformers for language
  understanding.
\newblock \emph{arXiv preprint arXiv:1810.04805}.

\bibitem[{Epstude and Roese(2008)}]{epstude2008functional}
Kai Epstude and Neal~J Roese. 2008.
\newblock The functional theory of counterfactual thinking.
\newblock \emph{Personality and Social Psychology Review}, 12(2):168--192.

\bibitem[{Goodman(1947)}]{goodman1947problem}
Nelson Goodman. 1947.
\newblock The problem of counterfactual conditionals.
\newblock \emph{The Journal of Philosophy}, 44(5):113--128.

\bibitem[{Hobbs(2005)}]{hobbs2005toward}
Jerry~R Hobbs. 2005.
\newblock Toward a useful concept of causality for lexical semantics.
\newblock \emph{Journal of Semantics}, 22(2):181--209.

\bibitem[{Hu et~al.(2019)Hu, Shi, Tan, Wang, Yang, Zhao, He, Qin, Wang
  et~al.}]{hu2019texar}
Zhiting Hu, Haoran Shi, Bowen Tan, Wentao Wang, Zichao Yang, Tiancheng Zhao,
  Junxian He, Lianhui Qin, Di~Wang, et~al. 2019.
\newblock Texar: A modularized, versatile, and extensible toolkit for text
  generation.
\newblock In \emph{ACL, System Demonstrations}.

\bibitem[{Hu et~al.(2017)Hu, Yang, Liang, Salakhutdinov, and
  Xing}]{hu2017toward}
Zhiting Hu, Zichao Yang, Xiaodan Liang, Ruslan Salakhutdinov, and Eric~P Xing.
  2017.
\newblock Toward controlled generation of text.
\newblock In \emph{ICML}.

\bibitem[{Kilickaya et~al.(2017)Kilickaya, Erdem, Ikizler-Cinbis, and
  Erdem}]{Kilickaya2017ReevaluatingAM}
Mert Kilickaya, Aykut Erdem, Nazli Ikizler-Cinbis, and Erkut Erdem. 2017.
\newblock Re-evaluating automatic metrics for image captioning.
\newblock In \emph{EACL}.

\bibitem[{Koncel-Kedziorski et~al.(2016)Koncel-Kedziorski, Konstas,
  Zettlemoyer, and Hajishirzi}]{KoncelKedziorski2016ATA}
Rik Koncel-Kedziorski, Ioannis Konstas, Luke~S. Zettlemoyer, and Hannaneh
  Hajishirzi. 2016.
\newblock A theme-rewriting approach for generating algebra word problems.
\newblock In \emph{EMNLP}.

\bibitem[{Kusner et~al.(2015)Kusner, Sun, Kolkin, and
  Weinberger}]{Kusner2015FromWE}
Matt~J. Kusner, Yu~Sun, Nicholas~I. Kolkin, and Kilian~Q. Weinberger. 2015.
\newblock From word embeddings to document distances.
\newblock In \emph{ICML}.

\bibitem[{Lample et~al.(2019)Lample, Subramanian, Smith, Denoyer, Ranzato, and
  Boureau}]{Lample2018MultipleAttributeTR}
Guillaume Lample, Sandeep Subramanian, Eric~Michael Smith, Ludovic Denoyer,
  Marc'Aurelio Ranzato, and Y-Lan Boureau. 2019.
\newblock Multiple-attribute text rewriting.
\newblock In \emph{ICLR}.

\bibitem[{Lawrence and Riezler(2018)}]{Lawrence2018ImprovingAN}
Carolin Lawrence and Stefan Riezler. 2018.
\newblock Improving a neural semantic parser by counterfactual learning from
  human bandit feedback.
\newblock In \emph{ACL}.

\bibitem[{Lin(2004)}]{lin2004rouge}
Chin-Yew Lin. 2004.
\newblock Rouge: A package for automatic evaluation of summaries.
\newblock \emph{Text Summarization Branches Out}.

\bibitem[{Liu et~al.(2016)Liu, Lowe, Serban, Noseworthy, Charlin, and
  Pineau}]{Liu2016HowNT}
Chia-Wei Liu, Ryan Lowe, Iulian Serban, Michael Noseworthy, Laurent Charlin,
  and Joelle Pineau. 2016.
\newblock How not to evaluate your dialogue system: An empirical study of
  unsupervised evaluation metrics for dialogue response generation.
\newblock In \emph{EMNLP}.

\bibitem[{Lowe et~al.(2017)Lowe, Noseworthy, Serban, Angelard-Gontier, Bengio,
  and Pineau}]{Lowe2017TowardsAA}
Ryan Lowe, Michael Noseworthy, Iulian Serban, Nicolas Angelard-Gontier, Yoshua
  Bengio, and Joelle Pineau. 2017.
\newblock Towards an automatic turing test: Learning to evaluate dialogue
  responses.
\newblock In \emph{ICLR}.

\bibitem[{Mostafazadeh et~al.(2016)Mostafazadeh, Chambers, He, Parikh, Batra,
  Vanderwende, Kohli, and Allen}]{mostafazadeh2016corpus}
Nasrin Mostafazadeh, Nathanael Chambers, Xiaodong He, Devi Parikh, Dhruv Batra,
  Lucy Vanderwende, Pushmeet Kohli, and James Allen. 2016.
\newblock A corpus and evaluation framework for deeper understanding of
  commonsense stories.
\newblock \emph{arXiv preprint arXiv:1604.01696}.

\bibitem[{Niven and Kao(2019)}]{niven2019probing}
Timothy Niven and Hung-Yu Kao. 2019.
\newblock Probing neural network comprehension of natural language arguments.
\newblock \emph{arXiv preprint arXiv:1907.07355}.

\bibitem[{Papineni et~al.(2002)Papineni, Roukos, Ward, and
  Zhu}]{papineni2002bleu}
Kishore Papineni, Salim Roukos, Todd Ward, and Wei-Jing Zhu. 2002.
\newblock {BLEU}: a method for automatic evaluation of machine translation.
\newblock In \emph{ACL}, pages 311--318.

\bibitem[{Pearl(2000)}]{pearl2000causality}
Judea Pearl. 2000.
\newblock \emph{Causality: models, reasoning and inference}, volume~29.
\newblock Springer.

\bibitem[{Pennington et~al.(2014)Pennington, Socher, and
  Manning}]{Pennington2014GloveGV}
Jeffrey Pennington, Richard Socher, and Christopher~D. Manning. 2014.
\newblock Glove: Global vectors for word representation.
\newblock In \emph{EMNLP}.

\bibitem[{Pichotta and Mooney(2014)}]{pichotta2014statistical}
Karl Pichotta and Raymond Mooney. 2014.
\newblock Statistical script learning with multi-argument events.
\newblock In \emph{EACL}, pages 220--229.

\bibitem[{Radford et~al.(2018)Radford, Narasimhan, Salimans, and
  Sutskever}]{radford2018improving}
Alec Radford, Karthik Narasimhan, Tim Salimans, and Ilya Sutskever. 2018.
\newblock Improving language understanding by generative pre-training.

\bibitem[{Radford et~al.(2019)Radford, Wu, Child, Luan, Amodei, and
  Sutskever}]{radford2019language}
Alec Radford, Jeffrey Wu, Rewon Child, David Luan, Dario Amodei, and Ilya
  Sutskever. 2019.
\newblock Language models are unsupervised multitask learners.
\newblock \emph{OpenAI Blog}, 1:8.

\bibitem[{Roemmele et~al.(2011)Roemmele, Bejan, and
  Gordon}]{Roemmele2011SemEval2012T7}
Melissa Roemmele, Cosmin~Adrian Bejan, and Andrew~S. Gordon. 2011.
\newblock Semeval-2012 task 7: Choice of plausible alternatives: An evaluation
  of commonsense causal reasoning.
\newblock In \emph{SemEval@NAACL-HLT}.

\bibitem[{Sap et~al.(2019)Sap, LeBras, Allaway, Bhagavatula, Lourie, Rashkin,
  Roof, Smith, and Choi}]{Sap2019ATOMICAA}
Maarten Sap, Ronan LeBras, Emily Allaway, Chandra Bhagavatula, Nicholas Lourie,
  Hannah Rashkin, Brendan Roof, Noah~A Smith, and Yejin Choi. 2019.
\newblock {ATOMIC}: An atlas of machine commonsense for if-then reasoning.
\newblock In \emph{AAAI}.

\bibitem[{Shen et~al.(2017)Shen, Lei, Barzilay, and Jaakkola}]{shen2017style}
Tianxiao Shen, Tao Lei, Regina Barzilay, and Tommi Jaakkola. 2017.
\newblock Style transfer from non-parallel text by cross-alignment.
\newblock In \emph{NeurIPS}.

\bibitem[{Son et~al.(2017)Son, Buffone, Raso, Larche, Janocko, Zembroski,
  Schwartz, and Ungar}]{Son2017RecognizingCT}
Youngseo Son, Anneke Buffone, Joe Raso, Allegra Larche, Anthony Janocko, Kevin
  Zembroski, H.~Andrew Schwartz, and Lyle~H. Ungar. 2017.
\newblock Recognizing counterfactual thinking in social media texts.
\newblock In \emph{ACL}.

\bibitem[{Wiseman et~al.(2017)Wiseman, Shieber, and
  Rush}]{Wiseman2017ChallengesID}
Sam Wiseman, Stuart~M. Shieber, and Alexander~M. Rush. 2017.
\newblock Challenges in data-to-document generation.
\newblock In \emph{EMNLP}.

\bibitem[{Woodward(2002)}]{woodward2002mechanism}
Jim Woodward. 2002.
\newblock What is a mechanism? a counterfactual account.
\newblock \emph{Philosophy of Science}, 69(S3):S366--S377.

\bibitem[{Zellers et~al.(2018)Zellers, Bisk, Schwartz, and
  Choi}]{Zellers2018SWAGAL}
Rowan Zellers, Yonatan Bisk, Roy Schwartz, and Yejin Choi. 2018.
\newblock Swag: A large-scale adversarial dataset for grounded commonsense
  inference.
\newblock In \emph{EMNLP}.

\bibitem[{Zellers et~al.(2019)Zellers, Holtzman, Bisk, Farhadi, and
  Choi}]{Zellers2019HellaSwagCA}
Rowan Zellers, Ari Holtzman, Yonatan Bisk, Ali Farhadi, and Yejin Choi. 2019.
\newblock Hella{S}wag: Can a machine really finish your sentence?
\newblock In \emph{ACL}.

\bibitem[{Zhang et~al.(2019)Zhang, Kishore, Wu, Weinberger, and
  Artzi}]{zhang2019bertscore}
Tianyi Zhang, Varsha Kishore, Felix Wu, Kilian~Q Weinberger, and Yoav Artzi.
  2019.
\newblock {BERTScore}: Evaluating text generation with bert.
\newblock \emph{arXiv preprint arXiv:1904.09675}.

\bibitem[{Zhu et~al.(2015)Zhu, Kiros, Zemel, Salakhutdinov, Urtasun, Torralba,
  and Fidler}]{Zhu2015AligningBA}
Yukun Zhu, Ryan Kiros, Richard~S. Zemel, Ruslan~R. Salakhutdinov, Raquel
  Urtasun, Antonio Torralba, and Sanja Fidler. 2015.
\newblock Aligning books and movies: Towards story-like visual explanations by
  watching movies and reading books.
\newblock \emph{ICCV}, pages 19--27.

\end{thebibliography}
\bibliographystyle{acl_natbib}

\appendix
\newpage
\hfill

\section{Crowdsourcing Details}
\label{app:data}

\paragraph{Quality Control} Since this is a creative annotation task for crowdworkers, rather than a tagging or selection task, we need two groups of crowdworkers for two separate steps: 1) workers to create a counterfactual alternatives for the storylines, 2) workers to create a new story ending that is coherent and logically consistent with the previous context that only changes the original story arc to regain narrative consistency.
Crowdworkers with more than 5000 HITs and at least a 99\% acceptance rate can take our qualification test, in which we require each crowdworker to do 3 HITs before being approved for the full task. We encourage workers to submit feedback to help us improve our instructions.

\paragraph{Cost} We pay \$0.24 to crowdworkers per instance for Step 1 and \$0.36 per instance for Step 2.

\section{Training Hyperparameters}
\label{app:train}

\paragraph{GPT2}
Text is encoded with BPE using a vocabulary size of 50,257. We set the maximum sequence length to 128 tokens, which we found is large enough to contain complete stories. We use Adam optimization with an initial learning rate of $10^{-5}$ and a minibatch size of 2. We train the models for 10K iterations using early stopping to select the model that does the best on the validation set. At inference time, we generate using the same procedure outlined in~\citet{radford2019language}: top-$k$ sampling with temperature set to 0.7 and $k$ set to 40.

\paragraph{GPT}
All models follow the setting of GPT \cite{radford2018improving} that used a 12-layer decoder-only transformer with masked self-attention heads. Text is encoded with BPE using a vocabulary size of 40,000. As above, we set the maximum sequence length to 128 tokens. We use Adam optimization with an initial learning rate of $6.25^{-5}$. We train the models for 10K iterations using early stopping to select the model that does the best on the validation set. We use the same generation procedure as for GPT2.

\end{document}